\documentclass[10pt,twocolumn,letterpaper]{article}

\ifdefined\CVPRREVIEW
  \usepackage[review]{cvpr_compat}
\else
  \usepackage[pagenumbers]{cvpr_compat}
\fi

\usepackage{microtype}
\usepackage{multirow}
\usepackage{array}
\usepackage{enumitem}
\usepackage{mathtools}
\usepackage{url}
\usepackage{ifthen}
\usepackage{placeins}
\usepackage{balance}
\definecolor{cvprblue}{rgb}{0.21,0.49,0.74}
\usepackage[pagebackref,breaklinks,colorlinks,allcolors=cvprblue]{hyperref}

\setlist[itemize]{leftmargin=*,topsep=2pt,itemsep=1pt,parsep=0pt}
\newcommand{\ER}{E_{\mathrm{R}}}
\newcommand{\HR}{H_{\mathrm{R}}}
\newcommand{\ES}{E_{\mathrm{S}}}
\newcommand{\HS}{H_{\mathrm{S}}}
\newcommand{\TaxSupCon}{\textsc{Taxonomy-SupCon}}

\title{What Drives Hierarchy-Aware Image Retrieval?\\
Taxonomy Alignment, Objective Choice, and Geometry}

\author{Ling Shi\\
Southeast University}

\begin{document}
\maketitle

\begin{abstract}
Foundation vision models provide strong generic representations, yet high class-level retrieval accuracy does not necessarily imply that an embedding respects a target semantic taxonomy. We study strict explicit-taxonomy image retrieval on frozen DINOv2 features and ask an attribution question: when hierarchical retrieval improves, how much of the change is associated with the organization of taxonomy-aware supervision, and how much with the Euclidean--hyperbolic geometry choice?

We evaluate higher levels with strict cross-class criteria that exclude finer-grained matches, and compare Euclidean and hyperbolic projections trained with either taxonomy-distance regression or a taxonomy-aware supervised contrastive objective. A compute-matched $2\times2$ Geometry $\times$ Loss factorial uses the same 768--256--32 projector capacity, optimization schedule, batch order, and fixed 100-epoch budget; here the Loss axis denotes the Regression-to-\TaxSupCon{} objective-family contrast. On CUB, the objective-family contrasts in mean hierarchy mAP---the average of strict middle- and high-level mAP, excluding Class/Leaf---are $+0.0487$ in Euclidean space and $+0.0414$ in hyperbolic space, compared with protocol-defined geometry contrasts of $+0.0102$ and $+0.0030$. On NABirds Parent-disjoint retrieval, the corresponding objective-family contrasts are $+0.0467$ and $+0.0440$, whereas the geometry contrasts are $+0.0017$ and $-0.0009$.

A complementary semantic-alignment control shows that the true taxonomy substantially outperforms a structure-preserving shuffled hierarchy, while a NABirds curvature/radius control does not support stronger negative curvature as the explanation for the observed hierarchy gains. Across the two evaluated taxonomies, the Regression-to-\TaxSupCon{} objective-family contrasts are larger in aggregate than the evaluated geometry contrasts; semantic alignment also matters in the separate control, while geometry remains hierarchy-dependent.
\end{abstract}

\section{Introduction}
\label{sec:intro}

Large-scale self-supervised vision models provide strong and transferable visual representations. DINOv2, for example, produces frozen features that perform well across recognition and retrieval tasks without downstream backbone fine-tuning~\cite{oquab2024dinov2}. However, visual discrimination and semantic organization are not identical objectives. In fine-grained retrieval, an embedding may reliably retrieve images from the same visual class while failing to preserve relations between \emph{different} classes that share a genus, family, Parent, or Supergroup.

This distinction matters when evaluating hierarchical representations. A same-class match is also trivially a same-Genus or same-Family match, so conventional higher-level retrieval can overestimate whether an embedding organizes different leaf classes according to the target taxonomy. We therefore use \textbf{strict cross-class criteria}: Cross-class Genus/Parent retrieval excludes the query leaf class, while Cross-genus Family and Cross-parent Supergroup additionally exclude the middle-level group. These criteria serve as an explicit-taxonomy test bed for attribution rather than as a claim that hierarchical image retrieval itself is new.

\begin{figure*}[t]
  \centering
  \includegraphics[width=\textwidth]{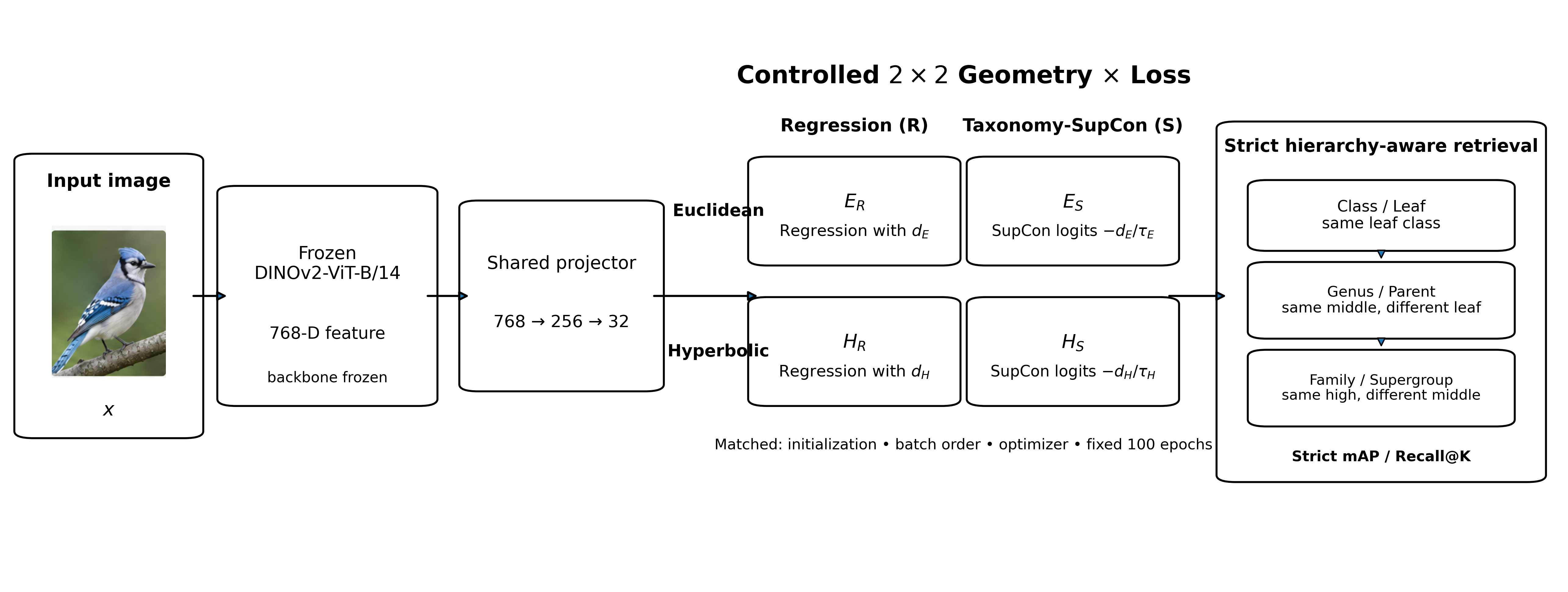}
  \caption{\textbf{Overview of the controlled hierarchy-aware retrieval framework.} A frozen DINOv2-ViT-B/14 encoder produces 768-D features, followed by a shared 768--256--32 projector family. We cross Euclidean/Hyperbolic geometry with taxonomy-distance Regression/\TaxSupCon{}, forming $\ER,\HR,\ES,\HS$ under matched initialization, batch order, optimizer, and fixed 100-epoch compute. Retrieval uses strict leaf-, middle-, and high-level positives that exclude finer-level matches.}
  \label{fig:framework}
\end{figure*}

Hyperbolic representation learning is a natural candidate because negatively curved spaces can compactly represent tree-like structures~\cite{nickel2017poincare,ganea2018hyperbolic}. Hyperbolic embeddings have been studied for retrieval, zero-shot recognition, metric learning, contrastive learning, and visual hierarchies~\cite{khrulkov2020hyperbolic,liu2020hyperbolic,ermolov2022hyperbolic,ge2023hyperbolic,wang2025visual}. Recent analysis further shows that temperature and hard-negative behavior can materially affect Euclidean--hyperbolic metric-learning comparisons~\cite{yue2024hardnegative}. These observations motivate separating geometry from objective design rather than interpreting any hyperbolic gain as a geometry-only effect.

We study this attribution problem on frozen DINOv2 features. On CUB~\cite{wah2011cub}, the target hierarchy is Class--Genus--Family; on NABirds~\cite{vanhorn2015nabirds}, it is Leaf--Parent--Supergroup. We distinguish three evidence questions: \textbf{(i) semantic alignment}---does the supervision correspond to the target taxonomy? \textbf{(ii) objective choice}---how do taxonomy-distance Regression and \TaxSupCon{} organize the same taxonomy relations? and \textbf{(iii) geometry}---what changes when Euclidean and Poincar\'e distances are compared under a matched training protocol?

For the latter two questions, we construct a compute-matched $2\times2$ Geometry $\times$ Loss factorial with Euclidean Regression ($\ER$), Hyperbolic Regression ($\HR$), Euclidean \TaxSupCon{} ($\ES$), and Hyperbolic \TaxSupCon{} ($\HS$). The Loss axis is a shorthand for this specific Regression-to-\TaxSupCon{} objective-family contrast, not an intervention on any single loss mechanism such as level balancing or negative handling. All cells use the same projector capacity, initialization, batch-order trajectory, optimizer, learning rate, and fixed epoch-100 endpoint. The Poincar\'e branch uses dataset-specific operating points frozen before the factorial comparison; thus our ``geometry effect'' is a protocol-defined Euclidean--Poincar\'e contrast, not a claim about a globally optimized geometry family or negative curvature in isolation.

On CUB, the Regression-to-\TaxSupCon{} objective-family contrast in mean hierarchy mAP is $+0.0487$ in Euclidean space and $+0.0414$ in hyperbolic space, while the geometry contrasts are $+0.0102$ and $+0.0030$. On NABirds Parent-disjoint retrieval, the corresponding objective-family contrasts are $+0.0467$ and $+0.0440$, whereas geometry contrasts shrink to $+0.0017$ and $-0.0009$. Under \TaxSupCon{}, the hyperbolic branch also shifts retrieval across levels: Genus decreases while Family increases on CUB, and Parent decreases while Supergroup increases on NABirds. We therefore interpret geometry as a smaller, level-dependent factor in the evaluated protocols rather than a stable global advantage.

Complementary controls answer different questions. A true-versus-structure-preserving-shuffle study provides semantic-alignment evidence, while a separate curvature/radius study bounds claims about negative-curvature mechanisms. We intentionally keep these controls distinct from the compute-matched factorial because they estimate different quantities.

\paragraph{Contributions.}
\begin{itemize}
  \item We formulate strict cross-class taxonomy retrieval as a controlled test bed that prevents finer-level matches from being re-counted as evidence of higher-level organization.
  \item We construct a compute-matched Geometry $\times$ Loss factorial that compares a specific Regression-to-\TaxSupCon{} objective-family contrast with a protocol-defined Euclidean--Poincar\'e contrast.
  \item Across CUB and NABirds Parent-disjoint, the loss-organization contrasts are substantially larger than the aggregate geometry contrasts, while the latter show a repeated hierarchy-level redistribution under \TaxSupCon{}.
  \item Separate semantic-alignment and mechanism-boundary controls show that performance is sensitive to the true taxonomy, while the NABirds curvature study does not support stronger negative curvature as the explanation for the gains.
\end{itemize}

\section{Related Work}
\label{sec:related}

\paragraph{Foundation visual representations.}
Self-supervised foundation models such as DINOv2 provide transferable features that can be frozen while downstream projections and objectives are changed~\cite{oquab2024dinov2}. We use this separation to study hierarchy-aware retrieval without conflating the result with backbone fine-tuning.

\paragraph{Hyperbolic metric and hierarchy learning.}
Poincar\'e embeddings and hyperbolic neural networks established hyperbolic representations for hierarchical data~\cite{nickel2017poincare,ganea2018hyperbolic}. Hyperbolic vision spans image embeddings, zero-shot recognition, metric learning, and contrastive representation learning~\cite{khrulkov2020hyperbolic,liu2020hyperbolic,ermolov2022hyperbolic,ge2023hyperbolic,mettes2024survey}. Closer to explicit visual hierarchies, Xu et al.~\cite{xu2023hiermargin} combine hyperbolic embeddings with hierarchical margins for coarse-to-fine recognition, Kwon et al.~\cite{kwon2024himapper} learn visual hierarchy mappings with a hierarchical contrastive loss, and Berg et al.~\cite{berg2025multiprototype} organize hyperbolic class prototypes using a known label hierarchy. These works exploit hierarchy to improve recognition or classification; our study instead asks how a specific objective-family contrast compares with a protocol-defined geometry contrast in strict retrieval. Yue et al.~\cite{yue2024hardnegative} analyze temperature and hard-negative effects in Euclidean--hyperbolic metric learning, underscoring the need to control the objective. HIER learns latent hyperbolic ancestor proxies without an explicit upper-level target taxonomy~\cite{kim2023hier}, whereas Wang et al.~\cite{wang2025visual} learn user-defined, part-based visual hierarchies without explicit hierarchical class labels and introduce distribution-based hierarchical retrieval evaluation. Our setting assumes an explicit taxonomic class tree and strict cross-class exclusions at each evaluated level.

\paragraph{Explicit hierarchy supervision.}
Taxonomy-aware metric learning predates recent contrastive objectives: Verma et al.~\cite{verma2012hierarchical} learn similarity metrics directly from a class taxonomy. Proxy Anchor and Supervised Contrastive Learning provide representative proxy- and sample-based metric objectives~\cite{kim2020proxy,khosla2020supcon}. Coarse-to-fine and hierarchical visual recognition explicitly model label granularity and taxonomic consistency~\cite{chang2021flamingo,park2025visually}; recent work additionally studies mixed-granularity supervision~\cite{park2026freegrained} and taxonomy-aware representation alignment~\cite{he2026tara}. Atigh et al.~\cite{atigh2026granularity} use hyperbolic prototypes for classification when labels may be provided at different hierarchy granularities. In contrast, our retrieval setting assumes a fully specified target taxonomy for training examples and asks which observed changes are associated with semantic alignment, objective choice, and the protocol-defined geometry contrast.

\section{Method}
\label{sec:method}

\subsection{Problem Formulation and Strict Retrieval}
Given image $x$, a frozen DINOv2-ViT-B/14 encoder produces $\mathbf f=F(x)\in\mathbb R^{768}$, which a trainable projector maps to a 32-D embedding. Each sample has leaf, middle, and high labels $y^c,y^m,y^h$: Class--Genus--Family on CUB and Leaf--Parent--Supergroup on NABirds.

We partition pairs into mutually exclusive relations
\begin{equation}
r(i,j)=
\begin{cases}
c,& y_i^c=y_j^c,\\
m,& y_i^m=y_j^m,\ y_i^c\neq y_j^c,\\
h,& y_i^h=y_j^h,\ y_i^m\neq y_j^m,\\
o,& y_i^h\neq y_j^h.
\end{cases}
\label{eq:relations}
\end{equation}
Strict middle-level positives share the middle group but differ in leaf class; strict high-level positives share the high group but differ at the middle level. Equivalently,
\begin{align}
\mathcal P_c(i)&=\{j:y_j^c=y_i^c,\ j\neq i\},\label{eq:Pc}\\
\mathcal P_m(i)&=\{j:y_j^m=y_i^m,\ y_j^c\neq y_i^c\},\label{eq:Pm}\\
\mathcal P_h(i)&=\{j:y_j^h=y_i^h,\ y_j^m\neq y_i^m\}.\label{eq:Ph}
\end{align}
The sets are mutually exclusive by construction. This yields Cross-class Genus / Cross-genus Family on CUB and Cross-class Parent / Cross-parent Supergroup on NABirds. Queries without a valid positive at the evaluated level are excluded. The same semantic partition is used by both training objectives, preventing an objective-specific definition of what constitutes a hierarchy relation.

\subsection{Euclidean and Hyperbolic Projectors}
Both geometries use the same 768--256--32 trainable capacity. The Euclidean branch outputs
\begin{equation}
\mathbf z_i^E=\frac{g_\theta(\mathbf f_i)}{\|g_\theta(\mathbf f_i)\|_2},
\end{equation}
and uses Euclidean distance, which is ranking-equivalent to cosine distance after normalization.

The hyperbolic branch follows the Poincar\'e-ball exponential-map construction used in hyperbolic neural networks~\cite{ganea2018hyperbolic}. It treats the MLP output as a tangent vector $\mathbf t_i$. Matching the implementation exactly, let
\begin{equation}
n_i=\max(\|\mathbf t_i\|_2,10^{-6}),\qquad
\mathbf t_i'=r\tanh(n_i)\frac{\mathbf t_i}{n_i},
\end{equation}
and map it to the Poincar\'e ball with the library exponential map and projection,
\begin{equation}
\mathbf z_i^H=\operatorname{proj}_{\mathbb D_c}\!\left(\exp_{\mathbf 0}^{c}(\mathbf t_i')\right).
\end{equation}
The dataset-specific operating points $(c,r)=(2.0,0.4)$ on CUB and $(0.1,0.4)$ on NABirds are frozen before the factorial final runs and are not retuned per cell or on official-test data. Throughout the paper, a \emph{geometry contrast} therefore denotes the protocol-defined Euclidean--Poincar\'e comparison at these frozen operating points; it is not a pure curvature intervention or a globally optimized comparison between geometry families.

\subsection{Taxonomy-Aware Objectives}
\paragraph{Distance regression.}
For every non-self pair, targets $\delta_{ij}\in\{0.2,0.5,0.8,1.0\}$ correspond respectively to relations $c,m,h,o$. For geometry $g\in\{E,H\}$,
\begin{equation}
\mathcal L_{\mathrm{reg}}^{(g)}
=
\frac{1}{|\Omega|}
\sum_{(i,j)\in\Omega}
\operatorname{SmoothL1}\!\left(d_g(i,j),\delta_{ij}\right).
\end{equation}
The controlled factorial deliberately uses the same target scale for both geometries so that the taxonomy supervision targets are held fixed while the distance parameterization changes. Consequently, the regression geometry contrast should be read as a matched-protocol contrast, not as a separately calibrated optimum for each geometry.

\paragraph{\TaxSupCon{}.}
Following the supervised-contrastive anchor/denominator construction~\cite{khosla2020supcon}, we replace a single positive set with mutually exclusive taxonomy levels. For each anchor, $\mathcal P_c(i),\mathcal P_m(i),\mathcal P_h(i)$ are the positive sets defined by Eqs.~\eqref{eq:Pc}--\eqref{eq:Ph}. We use geometry-matched distance logits
\begin{equation}
s_{ij}^{(g)}=-d_g(\mathbf z_i,\mathbf z_j)/\tau_g.
\end{equation}
Let $\mathcal A_\ell$ denote anchors with at least one positive at level $\ell$. Then
\begin{equation}
\mathcal L_\ell^{(g)}
=
-\frac{1}{|\mathcal A_\ell|}
\sum_{i\in\mathcal A_\ell}
\frac{1}{|\mathcal P_\ell(i)|}
\sum_{p\in\mathcal P_\ell(i)}
\log
\frac{e^{s_{ip}^{(g)}}}
{\sum_{a\neq i}e^{s_{ia}^{(g)}}}.
\end{equation}
Let $\mathcal V_B=\{\ell\in\{c,m,h\}:w_\ell>0,\ |\mathcal A_\ell|>0\}$ denote the hierarchy levels active in a minibatch. The optimized loss is
\begin{equation}
\mathcal L_{\mathrm{TS}}^{(g)}
=
\frac{\sum_{\ell\in\mathcal V_B}w_\ell\mathcal L_\ell^{(g)}}
{\sum_{\ell\in\mathcal V_B}w_\ell},
\qquad w_c=w_m=w_h=1.
\label{eq:taxsupcon_total}
\end{equation}
A level with no valid anchor is skipped and the remaining active weights are renormalized; a batch with no positive taxonomy relation at any active level is treated as an error. Samples outside the anchor's high-level group are not positive at any level but remain in the contrastive denominator. Thus each level is first averaged over valid anchors and only then combined, avoiding raw positive-pair-count weighting.

\subsection{Controlled Geometry $\times$ Loss Factorial}
The four cells are $\ER,\HR,\ES,\HS$. They share the frozen DINOv2 inputs, 768--256--32 projector capacity (205,600 trainable parameters), batch size 128, AdamW with learning rate $5\times10^{-5}$ and weight decay $10^{-4}$, dropout 0.1, gradient clipping at 1.0, and a fixed 100-epoch budget. Final training uses seeded example-level shuffling with \texttt{drop\_last=True}, \texttt{num\_workers=0}, and no class-balanced sampler, giving 46 updates/epoch on CUB and 148 on NABirds. Within each seed, all cells use the same trainable initialization and identical batch-order trajectory. Seeds are 42, 2024, and 3407.

\TaxSupCon{} temperatures are selected independently for Euclidean and hyperbolic branches using the same validation-only candidate set $\{0.07,0.10,0.20,0.30,0.50,1.00\}$ and the same fixed 50-epoch tuning budget. The selected values happen to coincide within each dataset: $\tau_E=\tau_H=0.20$ on CUB and $\tau_E=\tau_H=0.50$ on NABirds. The epoch-100 checkpoint is the fixed primary endpoint; validation-best checkpoints are retained only as auxiliary diagnostics and never replace the primary comparison. Official test data are not used to choose temperature, curvature, radius, or duration. This fixed-endpoint design is important because otherwise each factorial cell could receive a different effective optimization budget.

Rather than averaging across the other factor and reporting only canonical factorial main effects, we report protocol-specified \emph{simple effects}. This keeps the estimand explicit: the protocol-defined geometry contrast is measured separately under Regression and \TaxSupCon{}, while the Regression-to-\TaxSupCon{} objective-family contrast is measured separately in Euclidean and Poincar\'e space. For continuity with the factorial notation, we retain $\Delta_{\mathrm{loss}}$ as shorthand for this objective-family contrast rather than as a claim about a single loss mechanism. We report
\begin{align}
\Delta_{\mathrm{geom}}^R&=\HR-\ER,&
\Delta_{\mathrm{geom}}^S&=\HS-\ES,\\
\Delta_{\mathrm{loss}}^E&=\ES-\ER,&
\Delta_{\mathrm{loss}}^H&=\HS-\HR,
\end{align}
and interaction $\Delta_{\mathrm{int}}=(\HS-\ES)-(\HR-\ER)$. Three query/gallery perturbations are first averaged within each training seed.

\section{Experimental Setup}
\label{sec:setup}

\paragraph{Datasets and hierarchies.}
We evaluate on CUB-200-2011~\cite{wah2011cub} and NABirds~\cite{vanhorn2015nabirds}. CUB contains 11,788 images across 200 leaf classes; we train on the official 5,994-image training split and evaluate on the 5,794-image test split. Its Class--Genus--Family taxonomy is derived from the eBird/Clements Checklist v2025~\cite{clements2025checklist} and canonicalized to 124 genera and 37 families. NABirds uses its native Leaf--Parent--Supergroup hierarchy. The primary NABirds factorial is conducted on a Parent-disjoint protocol in which leaf classes and Parent groups are disjoint between development and test. The development bundle contains 19,021 images and the held-out test 5,017 images.

\paragraph{Validation and test isolation.}
The NABirds tuning split uses 13,959 training images, 4,049 gallery images, and 1,013 query images. Hyperparameters are selected only from the corresponding training-internal validation data. CUB and NABirds \TaxSupCon{} use the same six-temperature candidate set and fixed 50-epoch tuning budget described in Sec.~\ref{sec:method}. Official-test metrics are materialized only after the geometry operating points, temperatures, seeds, fixed epoch-100 endpoint, completed training runs, and pre-test fairness checks are frozen.

\paragraph{Evaluation and statistical unit.}
We report Recall@$K$ and mAP, using Euclidean/cosine-equivalent ranking for normalized Euclidean embeddings and Poincar\'e distance for hyperbolic embeddings. Each CUB query/gallery perturbation contains 870 query and 4,924 gallery images; the strict Genus and Family valid-query counts are approximately 482--487 and 747--750, respectively. CUB retains the registered query-micro factorial endpoint. NABirds Parent-disjoint official evaluation uses three fixed perturbations of 1,003 query and 4,014 gallery images. It uses group-macro aggregation: valid queries are first averaged within each evaluable Parent or Supergroup, after which groups receive equal weight. This prevents large groups from dominating the disjoint evaluation.

For every trained model, three fixed query/gallery perturbations are evaluated. These are \emph{not} treated as independent training observations: they are averaged within each training seed, and the three training seeds form the independent statistical units. Means and sample standard deviations therefore summarize $n=3$ independently trained models. We use seed directions and variability descriptively rather than as high-powered significance tests.

\paragraph{Contextual baselines and fairness checks.}
All learned methods operate on frozen DINOv2-ViT-B/14 features. We retain frozen DINOv2, PCA, a frozen-backbone HIER-style model, earlier matched-geometry controls, hierarchy-disjoint splits, semantic-supervision ablations, and curvature/radius controls as contextual or robustness evidence. They are not substituted for the compute-matched factorial when reporting the protocol-defined geometry and Regression-to-\TaxSupCon{} objective-family contrasts. Before official-test evaluation, the factorial audits verify 205,600 trainable parameters in every cell, matched trainable initialization within seed, identical batch-order trajectories and update counts, exact checkpoint reloads, and valid hyperbolic ball constraints. Full split statistics, implementation details, validation grids, factorial Recall@$K$, and audit metadata are provided in the supplement.

\section{Results}
\label{sec:results}

\subsection{Baseline Context}
Strong leaf-level retrieval does not guarantee taxonomy alignment. On CUB, frozen DINOv2 reaches Class mAP 0.6559 but strict Genus/Family mAP of only 0.3463/0.4000. HIER-style improves Class mAP to 0.7566 while its strict Genus/Family mAP falls to 0.2425/0.1696. A historical 32-D Euclidean taxonomy-regression projector instead obtains 0.4880/0.5631 strict Genus/Family mAP, illustrating that explicit target-taxonomy training can reorganize retrieval even when leaf-level performance decreases.

The same distinction appears on NABirds. In the original seen-class evaluation, HIER-style attains Class mAP 0.7031 but only 0.1149 strict Parent mAP and 0.3801 strict Supergroup mAP. The historical Taxonomy-SupCon baseline instead attains 0.3286 Parent mAP and 0.8901 Supergroup mAP while lowering Class mAP to 0.5249. These contextual comparisons show that target-taxonomy objectives can substantially change the semantic scale emphasized by retrieval. Because these models use different objectives and selection protocols, however, their cross-method gaps are not used as causal estimates of geometry or objective-family mechanisms.

\subsection{Objective-Family Contrasts Are Larger Than Geometry Contrasts}

\begin{table*}[t]
\centering
\caption{\textbf{Controlled Geometry $\times$ Loss factorial.} The Loss axis denotes the Regression-to-\TaxSupCon{} objective-family contrast. All cells share projector capacity, initialization within seed, batch sequence, optimizer, and fixed 100-epoch training. CUB reports its registered query-micro endpoint; NABirds Parent-disjoint reports group-macro. Mean hierarchy mAP averages strict middle and strict high mAP only (Class/Leaf excluded). Values are mean $\pm$ sample SD over three training seeds.}
\label{tab:factorial}
\setlength{\tabcolsep}{5.2pt}
\small
\begin{tabular}{llcccc}
\toprule
Dataset & Arm & Class mAP & Strict middle mAP & Strict high mAP & Mean hierarchy mAP\\
\midrule
\multirow{4}{*}{CUB}
& $\ER$ & $0.6171\pm0.0028$ & $0.4991\pm0.0017$ & $0.5636\pm0.0010$ & $0.5314\pm0.0013$\\
& $\HR$ & $0.6351\pm0.0033$ & $0.5153\pm0.0029$ & $0.5679\pm0.0006$ & $0.5416\pm0.0017$\\
& $\ES$ & $0.5144\pm0.0032$ & $0.5480\pm0.0022$ & $0.6121\pm0.0007$ & $0.5800\pm0.0009$\\
& $\HS$ & $0.4550\pm0.0016$ & $0.5405\pm0.0022$ & $0.6255\pm0.0006$ & $0.5830\pm0.0009$\\
\midrule
\multirow{4}{*}{NABirds}
& $\ER$ & $0.6831\pm0.0028$ & $0.1994\pm0.0034$ & $0.4830\pm0.0029$ & $0.3412\pm0.0031$\\
& $\HR$ & $0.6840\pm0.0052$ & $0.1971\pm0.0019$ & $0.4887\pm0.0024$ & $0.3429\pm0.0005$\\
& $\ES$ & $0.4655\pm0.0045$ & $0.2296\pm0.0027$ & $0.5461\pm0.0003$ & $0.3878\pm0.0014$\\
& $\HS$ & $0.4387\pm0.0066$ & $0.2245\pm0.0010$ & $0.5494\pm0.0031$ & $0.3869\pm0.0020$\\
\bottomrule
\end{tabular}
\end{table*}

On CUB, the mean-hierarchy objective-family contrasts are $\Delta_{\mathrm{loss}}^E=+0.048670\pm0.000409$ and $\Delta_{\mathrm{loss}}^H=+0.041424\pm0.001149$, both positive for all three training seeds. The corresponding protocol-defined geometry contrasts are smaller in observed magnitude: $\Delta_{\mathrm{geom}}^R=+0.010246\pm0.000416$ and $\Delta_{\mathrm{geom}}^S=+0.003001\pm0.000687$, again positive for all three seeds. The interaction is $-0.007245\pm0.000804$ and is negative in all three seeds, so the hyperbolic--Euclidean difference becomes smaller under the stronger taxonomy-aware objective.

NABirds Parent-disjoint shows the same ordering of observed magnitudes under a different hierarchy and unseen Parent groups. The objective-family contrasts are $\Delta_{\mathrm{loss}}^E=+0.046656\pm0.004351$ and $\Delta_{\mathrm{loss}}^H=+0.044046\pm0.002527$, positive in all three training seeds. By contrast, $\Delta_{\mathrm{geom}}^R=+0.001706\pm0.003657$ is positive in two of three seeds, while $\Delta_{\mathrm{geom}}^S=-0.000904\pm0.003188$ has a near-zero aggregate mean and mixed seed direction. The interaction is $-0.002610\pm0.006799$ and is not seed-wise consistent; we therefore do not interpret a negative interaction itself as replicated. The pattern repeated across the two evaluated datasets is narrower: the Regression-to-\TaxSupCon{} objective-family contrast is larger in aggregate observed magnitude than the protocol-defined geometry contrast under the evaluated operating regimes.

\begin{figure*}[t]
  \centering
  \includegraphics[width=.94\textwidth]{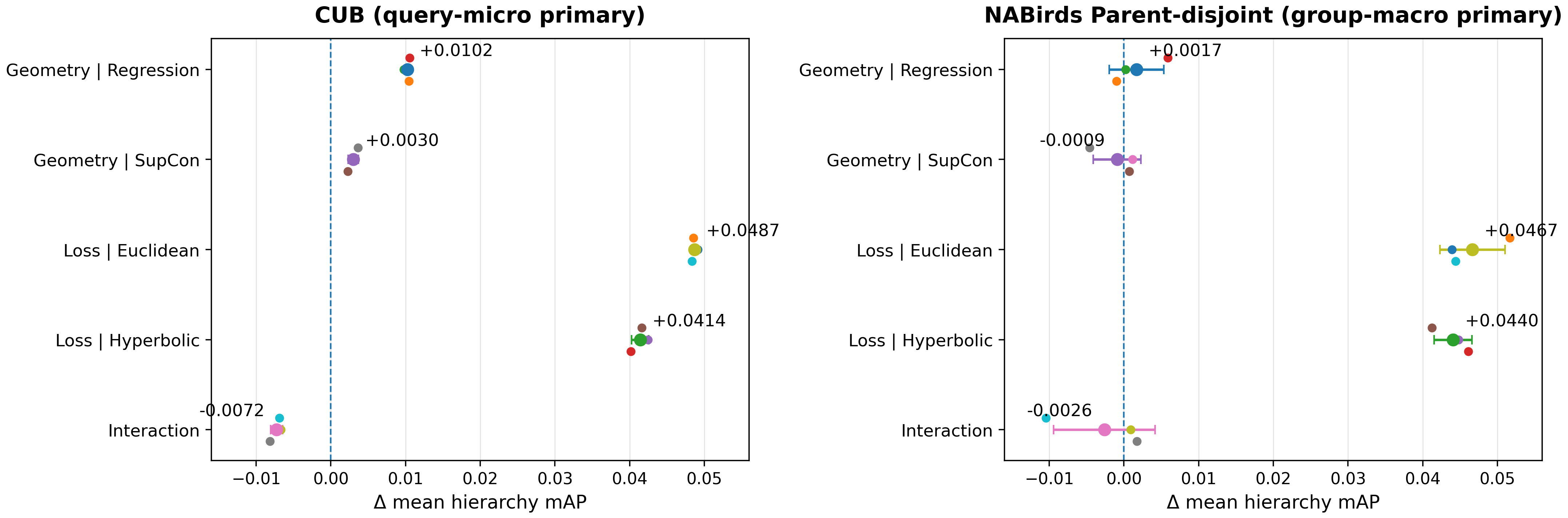}
  \caption{\textbf{Simple contrasts of objective family and geometry under matched compute.} The figure's ``Loss'' labels denote the Regression-to-\TaxSupCon{} objective-family contrast; $\Delta_{\mathrm{loss}}$ is retained only as shorthand notation. Each small point is one paired training-seed contrast; the larger point is the mean and the horizontal error bar is descriptive sample SD. Mean hierarchy mAP averages strict middle/high levels only and excludes Class/Leaf. CUB uses its query-micro primary endpoint and NABirds Parent-disjoint its group-macro primary endpoint.}
  \label{fig:effects}
\end{figure*}

For cross-dataset sensitivity, we additionally re-score the already frozen CUB projected-test embeddings using the same group-macro aggregation family as NABirds, without retraining or changing checkpoints. The resulting CUB effects are $+0.008904$ (Geometry under Regression), $+0.004647$ (Geometry under SupCon), $+0.062202$ (Loss under Euclidean), $+0.057945$ (Loss under Hyperbolic), and $-0.004256$ (interaction). Thus, the observed scale separation between the objective-family and geometry contrasts does not depend on using query-micro rather than group-macro on CUB. Because this re-scoring was performed after the primary experiment was frozen, it is a post-hoc sensitivity analysis and does not replace the registered CUB endpoint.

\subsection{Correct Semantic Alignment Matters}
The factorial compares two objectives that both use the correct taxonomy. We separately test whether the identity of the hierarchy matters.

Adding true Parent supervision raises mean hierarchy mAP from 0.3066 to 0.3424, a paired gain of $+0.0358$ that is positive for all three training seeds. Adding Supergroup supervision raises mean hierarchy mAP further to 0.3899, again positive for all three seeds. This latter gain is concentrated at the newly supervised high level: Supergroup mAP increases by $+0.0984$ relative to Leaf+Parent, whereas Parent mAP decreases by about 0.0034.

The structure-preserving shuffled taxonomy is a stronger semantic control. Full true-taxonomy supervision exceeds it by $+0.0849$ Parent mAP, $+0.2510$ Supergroup mAP, and $+0.1679$ mean hierarchy mAP, with all paired seed differences positive. Because the shuffle preserves the loss form and the multiset of hierarchy group sizes, the result supports the importance of semantic correspondence with the evaluation taxonomy rather than merely adding hierarchy-shaped contrastive terms. At the same time, Class mAP falls from 0.6717 under Leaf-only supervision to 0.5459 under the full hierarchy. Taxonomy-aware learning therefore induces a semantic-granularity trade-off rather than uniformly improving all retrieval levels.

These four supervision arms use independently validation-frozen training durations, so we treat the experiment as a semantic-alignment control rather than a compute-matched effect estimate. The main factorial, rather than this ablation, is the basis for comparing the specific objective-family contrast with the protocol-defined geometry contrast.

\subsection{Geometry Redistributes Retrieval Across Semantic Levels}
The small aggregate geometry effects hide a repeated level-wise pattern. Under \TaxSupCon{}, $\HS-\ES$ changes CUB Class/Genus/Family mAP by $-0.0594/-0.0074/+0.0134$, and NABirds Class/Parent/Supergroup mAP by $-0.0268/-0.0051/+0.0033$. Thus, the hyperbolic branch reduces finer and middle-level metrics while improving the highest evaluated level in both datasets.

\begin{figure*}[t]
  \centering
  \includegraphics[width=.94\textwidth]{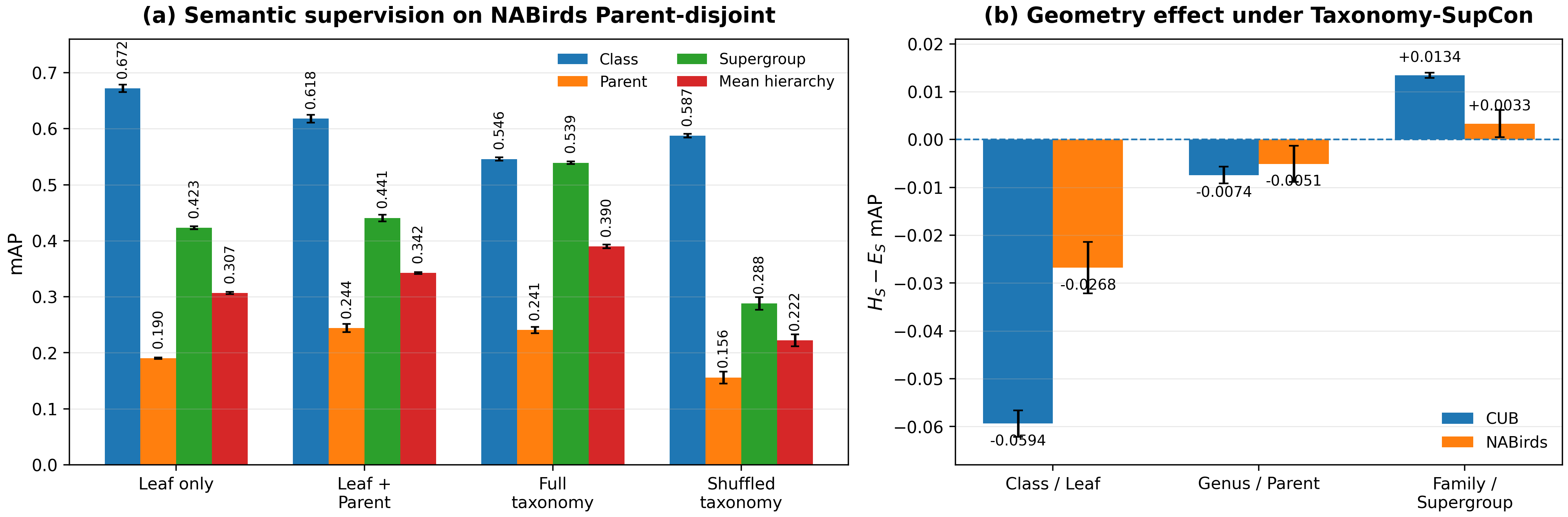}
  \caption{\textbf{Semantic supervision and hierarchy-level redistribution.} (a) NABirds Parent-disjoint mAP under Leaf-only, Leaf+Parent, full true-taxonomy, and structure-preserving shuffled-taxonomy supervision. (b) Per-level Hyperbolic$-$Euclidean effect under \TaxSupCon{}. In both datasets, the hyperbolic branch decreases finer/middle-level retrieval while improving the highest evaluated level.}
  \label{fig:tradeoff}
\end{figure*}

The repeated sign pattern is more informative than the small aggregate mean. In both datasets, the hyperbolic SupCon branch trades finer/middle-level retrieval for the highest evaluated level. We therefore describe the result as \emph{hierarchy-level redistribution}, not as evidence for a universal hyperbolic improvement. Nor does the factorial identify a radial-depth mechanism: it reports retrieval behavior at frozen Poincar'e operating points, while the separate mechanism study below is needed to bound curvature-based interpretations.

\subsection{Robustness and Mechanism Boundaries}
Across three CUB Genus-disjoint outer splits, mean hierarchy group-macro mAP is $0.3146\pm0.0098$ for Frozen DINOv2, $0.3126\pm0.0099$ for matched Euclidean, $0.3148\pm0.0096$ for matched Hyperbolic, and $0.3464\pm0.0218$ for Taxonomy-SupCon. The Taxonomy-SupCon--matched-Hyperbolic paired gain averages $+0.0316$ and is positive on all three splits, but its decomposition is strongly level-dependent: Genus mAP is lower by about 0.0003 while Family mAP is higher by about 0.0634.

Across three NABirds Parent-disjoint outer taxonomy splits, mean hierarchy mAP is $0.3404\pm0.0220$ for matched Euclidean, $0.3426\pm0.0223$ for matched Hyperbolic, and $0.3849\pm0.0177$ for Taxonomy-SupCon. Taxonomy-SupCon exceeds matched Hyperbolic by $0.0423\pm0.0068$, positive on every outer split, whereas matched Hyperbolic exceeds matched Euclidean by only $0.00222\pm0.00049$. These earlier experiments use validation-frozen configurations rather than the new fixed-100-epoch factorial, so they provide qualitative robustness rather than additional factorial estimates.

The curvature control further narrows the mechanism claim. At fixed $r=0.4$, $c=0.1$ is slightly below the near-flat $c=10^{-6}$ control in mean hierarchy mAP ($-0.000419$), and increasing curvature to $c=1.0$ reduces it by a further $0.002314$. Radius changes are larger but level-dependent. The factorial therefore measures a specific Poincar\'e parameterization at a frozen operating point; it does not establish stronger negative curvature as the cause of the hierarchy gains.

\section{Discussion and Limitations}
\label{sec:discussion}

\paragraph{Three evidence layers.}
The experiments separate evidence questions that are often changed together, but the two factorial axes should be interpreted at the level actually intervened on. The compute-matched factorial compares a \emph{Regression-to-\TaxSupCon{} objective-family contrast} with a \emph{protocol-defined geometry contrast} (Euclidean versus Poincar\'e at frozen operating points). The objective-family intervention jointly changes the objective form, level-wise aggregation, pair weighting, and negative handling; the factorial does not attribute its contrast to any one of those mechanisms. Across both datasets, this objective-family contrast is larger in aggregate observed magnitude than the evaluated Euclidean--Poincar\'e contrast.

\begin{table}[!t]
\centering
\caption{\textbf{Semantic-supervision ablation on NABirds Parent-disjoint.} Parent/Supergroup use group-macro mAP. Each arm uses its validation-frozen training duration; this tests semantic alignment rather than the compute-matched Regression-to-\TaxSupCon{} objective-family contrast.}
\label{tab:semantic}
\footnotesize
\setlength{\tabcolsep}{3.0pt}
\begin{tabular}{lcccc}
\toprule
Supervision & Class & Parent & Super. & Mean hier.\\
\midrule
Leaf only & .6717 & .1901 & .4231 & .3066\\
Leaf + Parent & .6178 & .2441 & .4407 & .3424\\
Full taxonomy & .5459 & .2408 & .5391 & .3899\\
Shuffled taxonomy & .5874 & .1559 & .2881 & .2220\\
\bottomrule
\end{tabular}
\end{table}

The true-versus-shuffled experiment addresses a different question. Both factorial losses already use the correct taxonomy, so the factorial cannot establish whether semantic identity matters. Disrupting that correspondence while preserving a hierarchy-shaped contrastive structure substantially reduces strict hierarchy retrieval. Because these supervision arms use independently validation-selected durations, we treat this as a \emph{semantic-alignment control}, not as another compute-matched factorial contrast.

\paragraph{What the geometry contrast means.}
The geometry contrasts should not be read as pure curvature effects or as a search over the best possible Euclidean and hyperbolic models. CUB and NABirds use dataset-specific Poincar\'e operating points frozen before the factorial final runs. Regression deliberately shares one absolute taxonomy-target scale across geometries, while \TaxSupCon{} uses a symmetric validation-only temperature search for both branches. These choices define a matched protocol in which supervision, capacity, update budget, initialization, and batch order are controlled, but they do not constitute independent global optimization of each geometry family.

The remaining geometry differences are also level-dependent. Under \TaxSupCon{}, the hyperbolic branch lowers Class and Genus while raising Family mAP on CUB, and lowers Class and Parent while raising Supergroup mAP on NABirds. The NABirds curvature/radius control further provides \emph{mechanism-boundary evidence}: the selected $c=0.1$ point does not outperform a near-flat control at fixed radius, and stronger curvature decreases mean hierarchy mAP. We therefore do not attribute the factorial geometry contrast specifically to negative curvature or radial depth.

\paragraph{Semantic trade-offs and robustness.}
The supervision ablation and factorial both show that improving strict hierarchy retrieval can reduce leaf-level discrimination. \TaxSupCon{} is therefore better aligned with the strict hierarchy objective studied here rather than universally superior as an embedding loss. Likewise, outer Genus/Parent-disjoint studies support the broader scale separation between taxonomy-aware objective changes and historical Euclidean--hyperbolic differences, but CUB also shows that an aggregate gain can be concentrated in the still-shared Family level. We treat these studies as robustness evidence, not additional factorial estimates or full subtree-disjoint generalization.

\paragraph{Scope and statistical limitations.}
All experiments use frozen DINOv2-ViT-B/14 features, a compact 32-D projector, and two related fine-grained bird datasets. The primary factorial contains three independent training seeds; query/gallery perturbations are averaged within seed, so sample standard deviations and seed directions are descriptive and we make no significance claims. The geometry contrast represents specific frozen operating points, and an equivalent full curvature/radius study has not been performed on CUB. The semantic shuffle uses one fixed structure-preserving permutation and is not compute-matched, supporting ``true taxonomy versus this controlled shuffle'' rather than a population-average random-taxonomy effect. CUB and NABirds also use different registered primary aggregation rules; the post-hoc CUB group-macro sensitivity preserves the qualitative separation but does not replace the query-micro primary endpoint.

Within these boundaries, the strongest repeated aggregate changes arise from the Regression-to-\TaxSupCon{} objective-family contrast, semantic-alignment evidence independently shows sensitivity to the target taxonomy, and the evaluated geometry contrast is smaller in aggregate while still redistributing retrieval across hierarchy levels.

\section{Conclusion}
\label{sec:conclusion}

We studied strict explicit-taxonomy image retrieval on frozen DINOv2 features and separated three evidence questions that are often conflated: semantic alignment with the target taxonomy, the choice between taxonomy-distance Regression and \TaxSupCon{}, and a protocol-defined Euclidean--Poincar\'e contrast. Across CUB and NABirds, the compute-matched Regression-to-\TaxSupCon{} objective-family contrasts are larger in aggregate observed magnitude than the evaluated geometry contrasts. A separate true-versus-shuffle control shows strong sensitivity to semantic alignment, while the NABirds curvature control does not support stronger negative curvature as the explanation for the observed hierarchy gains. In the evaluated settings, geometry remains relevant mainly through smaller, level-dependent redistribution rather than a stable overall advantage.
\label{body:end}

\end{document}


\maketitle

\section{Dataset and Taxonomy Construction}

\paragraph{CUB.}
CUB-200-2011 contains 11,788 images from 200 leaf classes, with the official 5,994/5,794 train/test split. The accepted taxonomy-construction record identifies the external source as the eBird/Clements Checklist v2025. CUB class names are converted to common names and matched to the checklist after lowercasing and removing punctuation, whitespace, hyphens, and underscores. Of the 200 classes, 163 are exact normalized-name matches, 18 use fuzzy matching, 18 use manually recorded aliases, and Yellow Warbler is completed by manual verification. Genus is taken from the first token of the matched scientific name. Family strings are canonicalized by stripping a trailing parenthetical common-name gloss; in particular, \texttt{Parulidae} and \texttt{Parulidae (New World Warblers)} are merged. The resulting hierarchy contains 200 classes, 124 genera, and 37 canonical families, with zero class-to-genus, class-to-family, or genus-to-family consistency conflicts. Eighty-seven of the 124 genera contain only one CUB class.

The archived source bundle contains the exact construction and canonicalization utilities, which are now included with the paper's reproducibility artifacts. It does not contain the external checklist CSV, the realized class-to-Genus-to-Family mapping CSV, or the alias JSON used for manual aliases. Exact taxonomy reconstruction therefore still requires those missing data artifacts.

\paragraph{NABirds.}
NABirds is used with its native tree rather than biological genus/family labels. Leaf is the visual leaf class, Parent is its direct parent node, and Supergroup is the first child below the root Birds. The processed hierarchy contains 555 leaf classes, 404 Parents, and 22 Supergroups. The primary factorial uses the fixed Parent-disjoint protocol: train/validation/test contain 328/114/113 leaf classes and 242/81/81 Parent groups, with zero pairwise leaf-class or Parent overlap. Train and validation together form the 19,021-sample development bundle; the held-out official test contains 5,017 images. Temperature tuning uses 13,959 training images plus a fixed 4,049-gallery/1,013-query validation split. Official testing uses 4,014 gallery and 1,003 query images per perturbation.

\section{Strict Retrieval Definitions and Aggregation}

For a query $q$ and gallery item $j$, let $y^c,y^m,y^h$ denote leaf, middle, and high-level labels. The strict positive masks are
\begin{align}
P_c(q,j) &= [y_q^c=y_j^c],\\
P_m(q,j) &= [y_q^m=y_j^m]\,[y_q^c\neq y_j^c],\\
P_h(q,j) &= [y_q^h=y_j^h]\,[y_q^m\neq y_j^m].
\end{align}
Thus CUB evaluates Class, Cross-class Genus, and Cross-genus Family; NABirds evaluates Leaf/Class, Cross-class Parent, and Cross-parent Supergroup. A query is valid for a level only when at least one corresponding positive exists in the gallery.

For a valid query $q$, Recall@$K$ is one if at least one positive occurs among the top $K$ retrieved items and zero otherwise. If $r_{qk}\in\{0,1\}$ indicates whether rank $k$ is positive and $N_q^+=\sum_k r_{qk}$, then
\begin{equation}
\mathrm{AP}(q)=\frac{1}{N_q^+}\sum_k
\left(\frac{\sum_{t\le k}r_{qt}}{k}\right) r_{qk}.
\end{equation}
Query-micro mAP averages AP equally over valid queries. For group-macro evaluation, valid query metrics are first averaged within each evaluable middle/high group and those group means are then averaged with equal weight. CUB's registered factorial endpoint is query-micro; NABirds Parent-disjoint uses group-macro. The post-hoc CUB group-macro analysis in Sec.~S11 does not replace the registered endpoint.

\section{Query/Gallery Perturbations and Statistical Unit}

The fixed query/gallery perturbation seeds are 42, 2024, and 3407. Every CUB perturbation contains 870 queries and 4,924 gallery images. Strict Genus valid-query counts are 487/482/483 and strict Family counts are 747/750/747 for Q/G seeds 42/2024/3407. Every NABirds Parent-disjoint perturbation contains 1,003 queries and 4,014 gallery images; strict Parent valid-query counts are 494/495/494 and strict Supergroup counts are 962/963/962, with 27 and 11 evaluable groups respectively.

For each trained model, the three Q/G perturbations are evaluated on the same projected official-test embedding. Metrics are first averaged across the three Q/G perturbations \emph{within} a training seed; only then are the three training seeds treated as independent statistical units. Hence all sample standard deviations for learned factorial models use $n=3$ training seeds rather than nine Q/G cells.

The accepted final audits record a SHA256 identifier for each fixed Q/G split and verify the same identifiers across all factorial arms. Phase 6C-4 recovered the project's generic retrieval-split utility and the NABirds Parent-disjoint preparation/split utility from the archived source bundle, together with a manifest of the accepted final split hashes and sizes. The final Q/G NPZ index arrays themselves are not present, so the accepted hashes remain the authoritative split identifiers rather than newly regenerated index lists.

\section{Full Factorial Implementation Details}

Both geometries use the same trainable MLP:
\[
768 \rightarrow \mathrm{Linear}(256)\rightarrow\mathrm{BN}\rightarrow
\mathrm{ReLU}\rightarrow\mathrm{Dropout}(0.1)\rightarrow\mathrm{Linear}(32),
\]
for 205,600 trainable parameters. Euclidean outputs are $\ell_2$ normalized. Hyperbolic outputs use the mapping in Sec.~S5.

Final factorial training uses AdamW with learning rate $5\times10^{-5}$, constant learning-rate schedule, weight decay $10^{-4}$, batch size 128, gradient clipping at norm 1.0, \texttt{num\_workers=0}, and \texttt{drop\_last=True}. The loader uses ordinary example-level random shuffling (\texttt{shuffle=True}) with a PyTorch generator seeded by the training seed; no class-balanced or hierarchy-balanced sampler is used. CUB trains on all 5,994 training examples and therefore executes 46 updates per epoch; NABirds trains on all 19,021 development examples and executes 148 updates per epoch. Every final cell is trained for exactly 100 epochs.

For each training seed (42, 2024, 3407), the random state is reset before model construction. The accepted pre-test audits verify identical trainable initialization hashes, identical epoch-wise batch-order hashes, identical update counts, and equal parameter counts across $\ER,\HR,\ES,\HS$ within that seed. Each final checkpoint is reloaded into a fresh model and reproduces the projected verification sample with maximum absolute difference 0.

Regression minimizes SmoothL1 over every non-self pair using the shared targets
\[
\delta_c=0.2,\quad \delta_m=0.5,\quad \delta_h=0.8,\quad \delta_o=1.0.
\]
The target scale is intentionally identical for Euclidean and hyperbolic branches so that the taxonomy supervision targets remain fixed while the distance parameterization changes. It should therefore be read as a protocol-defined matched contrast rather than separately calibrated geometry optima.

\paragraph{\TaxonomySupCon{} implementation.}
For $\ell\in\{c,m,h\}$, let $\mathcal P_\ell(i)$ be the mutually exclusive positive sets in the main paper and $\mathcal A_\ell=\{i:|\mathcal P_\ell(i)|>0\}$. With $s_{ij}=-d_g(z_i,z_j)/\tau_g$,
\[
\mathcal L_\ell =
-\frac{1}{|\mathcal A_\ell|}\sum_{i\in\mathcal A_\ell}
\frac{1}{|\mathcal P_\ell(i)|}\sum_{p\in\mathcal P_\ell(i)}
\log\frac{e^{s_{ip}}}{\sum_{a\ne i} e^{s_{ia}}}.
\]
Let $\mathcal V_B=\{\ell:w_\ell>0,\ |\mathcal A_\ell|>0\}$ for the current minibatch. The implemented total objective is
\[
\mathcal L_{\rm TS} =
\frac{\sum_{\ell\in\mathcal V_B}w_\ell\mathcal L_\ell}
{\sum_{\ell\in\mathcal V_B}w_\ell},
\qquad
w_c=w_m=w_h=1.
\]
If one level has no valid anchor, that level is skipped and the remaining active weights are renormalized. A minibatch with no positive taxonomy relation at any active level raises an error. All non-self examples, including samples outside the anchor's high-level group, remain in the contrastive denominator.

\section{Hyperbolic Implementation and Operating Points}

The implementation instantiates \texttt{geoopt.PoincareBall(c)} with $c>0$ representing absolute curvature, i.e. sectional curvature $-c$, and mathematical ball
\[
\mathbb D_c^d=\{x\in\mathbb R^d:c\|x\|_2^2<1\},
\qquad R_c=\frac{1}{\sqrt c}.
\]
For the MLP output $t$, the project code uses
\[
n=\max(\|t\|_2,10^{-6}),\qquad
t'=r\tanh(n)\frac{t}{n},
\]
then calls \texttt{ball.expmap0(t')} followed by \texttt{ball.projx}. Under the corresponding Poincar\'e-ball convention,
\[
\exp_0^c(v)=
\tanh(\sqrt c\,\|v\|_2)\frac{v}{\sqrt c\,\|v\|_2}
\]
with the continuous value at $v=0$. The distance used by the code is \texttt{ball.dist}; equivalently,
\[
d_c(x,y)=\frac{2}{\sqrt c}\operatorname{artanh}
\!\left(\sqrt c\,\|(-x)\oplus_c y\|_2\right),
\]
where
\[
x\oplus_c y=
\frac{(1+2c\langle x,y\rangle+c\|y\|^2)x+
(1-c\|x\|^2)y}
{1+2c\langle x,y\rangle+c^2\|x\|^2\|y\|^2}.
\]
Before hyperbolic pairwise distances, embeddings are projected with \texttt{ball.projx}. During training, non-finite distances are sanitized and the distance matrix is clipped to $[0,20]$; official-test ranking uses the same library distance with sanitization and an upper clipping bound of $10^6$.

The CUB operating point is $(c,r)=(2.0,0.4)$, giving ball radius $1/\sqrt2=0.7071$; the NABirds point is $(0.1,0.4)$, giving radius $3.1623$. Protocol locks state that each is inherited from a prior dataset-specific operating point and frozen before the factorial final runs; neither is retuned per factorial cell or on official-test data. Across all accepted hyperbolic factorial runs, the maximum audited embedding norm is 0.3622 on CUB and 0.3979 on NABirds, safely inside the respective balls.

The archived project code does not set a manual boundary epsilon for \texttt{projx} and the accepted artifacts do not pin the Geoopt package version. Therefore we do not invent a library-internal projection margin; the reproducible project-level numerical constant is the $10^{-6}$ tangent-norm clamp above.

\section{Validation-Only Temperature Selection}

Euclidean and hyperbolic \TaxonomySupCon{} branches are tuned independently with the same six candidates, the same seed 42, and the same fixed 50-epoch budget. Candidate checkpoints are evaluated only at epoch 50 and are not reused for final training. All final cells are retrained from scratch for 100 epochs after temperature selection.

\begin{table}[h]
\centering
\caption{CUB temperature selection. The score is the query-micro mean of strict Genus and strict Family mAP.}
\small
\begin{tabular}{llrrrr}
\toprule
Geometry & $\tau$ & Genus mAP & Family mAP & Mean & Status\\
\midrule
Euclidean & 0.07 & 0.537612 & 0.617534 & 0.577573 & \\
Euclidean & 0.10 & 0.541775 & 0.618417 & 0.580096 & \\
Euclidean & 0.20 & 0.549530 & 0.629024 & 0.589277 & winner\\
Euclidean & 0.30 & 0.518694 & 0.658657 & 0.588676 & runner-up\\
Euclidean & 0.50 & 0.424330 & 0.696066 & 0.560198 & \\
Euclidean & 1.00 & 0.348874 & 0.712698 & 0.530786 & \\
Hyperbolic & 0.07 & 0.537128 & 0.618151 & 0.577640 & \\
Hyperbolic & 0.10 & 0.536336 & 0.621024 & 0.578680 & \\
Hyperbolic & 0.20 & 0.540021 & 0.641901 & 0.590961 & winner\\
Hyperbolic & 0.30 & 0.486798 & 0.674783 & 0.580791 & runner-up\\
Hyperbolic & 0.50 & 0.393841 & 0.703824 & 0.548832 & \\
Hyperbolic & 1.00 & 0.334421 & 0.711251 & 0.522836 & \\
\bottomrule
\end{tabular}
\end{table}

CUB selects $\tau_E=\tau_H=0.20$. The Euclidean and hyperbolic runner-up is $\tau=0.30$.

\begin{table}[h]
\centering
\caption{NABirds Parent-disjoint temperature selection. Parent and Supergroup are group-macro mAP; the score is their mean.}
\small
\begin{tabular}{llrrrr}
\toprule
Geometry & $\tau$ & Parent mAP & Supergroup mAP & Mean & Status\\
\midrule
Euclidean & 0.07 & 0.228385 & 0.544684 & 0.386535 & \\
Euclidean & 0.10 & 0.229470 & 0.547203 & 0.388336 & \\
Euclidean & 0.20 & 0.232159 & 0.557994 & 0.395076 & \\
Euclidean & 0.30 & 0.219048 & 0.574292 & 0.396670 & runner-up\\
Euclidean & 0.50 & 0.201959 & 0.595510 & 0.398735 & winner\\
Euclidean & 1.00 & 0.180174 & 0.603103 & 0.391639 & \\
Hyperbolic & 0.07 & 0.229017 & 0.546425 & 0.387721 & \\
Hyperbolic & 0.10 & 0.230315 & 0.550430 & 0.390373 & \\
Hyperbolic & 0.20 & 0.225752 & 0.565639 & 0.395696 & \\
Hyperbolic & 0.30 & 0.210454 & 0.585483 & 0.397969 & runner-up\\
Hyperbolic & 0.50 & 0.197504 & 0.598966 & 0.398235 & winner\\
Hyperbolic & 1.00 & 0.169683 & 0.603121 & 0.386402 & \\
\bottomrule
\end{tabular}
\end{table}

NABirds selects $\tau_E=\tau_H=0.50$, with $\tau=0.30$ as the runner-up for both geometries. The accepted temperature audits record \texttt{official\_test\_accessed=false}, common trainable initialization, identical batch-order trajectories, and matched parameter counts across candidates.

\section{Seed-Level Geometry $\times$ Loss Factorial}

\begin{table}[h]
\centering
\caption{CUB registered query-micro mAP after averaging the three Q/G perturbations within each training seed.}
\small
\begin{tabular}{llrrrr}
\toprule
Arm & Seed & Class & Genus & Family & Mean hierarchy\\
\midrule
$E_R$ & 42 & 0.613884 & 0.500621 & 0.564324 & 0.532472\\
$E_R$ & 2024 & 0.618459 & 0.497207 & 0.562511 & 0.529859\\
$E_R$ & 3407 & 0.619024 & 0.499484 & 0.563990 & 0.531737\\
$H_R$ & 42 & 0.632376 & 0.517540 & 0.568281 & 0.542911\\
$H_R$ & 2024 & 0.638752 & 0.512089 & 0.567167 & 0.539628\\
$H_R$ & 3407 & 0.634214 & 0.516354 & 0.568181 & 0.542268\\
$E_S$ & 42 & 0.514624 & 0.550282 & 0.611362 & 0.580822\\
$E_S$ & 2024 & 0.517443 & 0.545790 & 0.612189 & 0.578990\\
$E_S$ & 3407 & 0.511104 & 0.547848 & 0.612682 & 0.580265\\
$H_S$ & 42 & 0.456725 & 0.541272 & 0.624918 & 0.583095\\
$H_S$ & 2024 & 0.454890 & 0.538037 & 0.626124 & 0.582080\\
$H_S$ & 3407 & 0.453435 & 0.542295 & 0.625512 & 0.583904\\
\bottomrule
\end{tabular}
\end{table}

\begin{table}[h]
\centering
\caption{CUB mean-hierarchy simple effects by independent training seed. Final column is mean $\pm$ sample SD.}
\small
\begin{tabular}{lrrrr}
\toprule
Effect & 42 & 2024 & 3407 & Mean $\pm$ SD\\
\midrule
$\Delta_{\rm geom}^{R}$ & +0.010438 & +0.009769 & +0.010531 & $+0.010246\pm0.000416$\\
$\Delta_{\rm geom}^{S}$ & +0.002273 & +0.003090 & +0.003638 & $+0.003001\pm0.000687$\\
$\Delta_{\rm loss}^{E}$ & +0.048350 & +0.049131 & +0.048529 & $+0.048670\pm0.000409$\\
$\Delta_{\rm loss}^{H}$ & +0.040184 & +0.042452 & +0.041636 & $+0.041424\pm0.001149$\\
$\Delta_{\rm int}$ & -0.008165 & -0.006678 & -0.006892 & $-0.007245\pm0.000804$\\
\bottomrule
\end{tabular}
\end{table}

\begin{table}[h]
\centering
\caption{NABirds Parent-disjoint group-macro mAP after averaging Q/G perturbations within training seed.}
\small
\begin{tabular}{llrrrr}
\toprule
Arm & Seed & Class & Parent & Supergroup & Mean hierarchy\\
\midrule
$E_R$ & 42 & 0.684085 & 0.201829 & 0.484932 & 0.343381\\
$E_R$ & 2024 & 0.685218 & 0.200846 & 0.484305 & 0.342575\\
$E_R$ & 3407 & 0.679945 & 0.195478 & 0.479702 & 0.337590\\
$H_R$ & 42 & 0.685569 & 0.196332 & 0.488446 & 0.342389\\
$H_R$ & 2024 & 0.688281 & 0.199166 & 0.486469 & 0.342818\\
$H_R$ & 3407 & 0.678195 & 0.195661 & 0.491257 & 0.343459\\
$E_S$ & 42 & 0.460400 & 0.229066 & 0.546453 & 0.387759\\
$E_S$ & 2024 & 0.467234 & 0.227171 & 0.545812 & 0.386491\\
$E_S$ & 3407 & 0.468878 & 0.232590 & 0.545936 & 0.389263\\
$H_S$ & 42 & 0.435353 & 0.224894 & 0.552094 & 0.388494\\
$H_S$ & 2024 & 0.434436 & 0.225294 & 0.549954 & 0.387624\\
$H_S$ & 3407 & 0.446344 & 0.223310 & 0.546061 & 0.384685\\
\bottomrule
\end{tabular}
\end{table}

\begin{table}[h]
\centering
\caption{NABirds mean-hierarchy group-macro simple effects by independent training seed.}
\small
\begin{tabular}{lrrrr}
\toprule
Effect & 42 & 2024 & 3407 & Mean $\pm$ SD\\
\midrule
$\Delta_{\rm geom}^{R}$ & -0.000992 & +0.000242 & +0.005869 & $+0.001706\pm0.003657$\\
$\Delta_{\rm geom}^{S}$ & +0.000734 & +0.001133 & -0.004577 & $-0.000904\pm0.003188$\\
$\Delta_{\rm loss}^{E}$ & +0.044379 & +0.043916 & +0.051673 & $+0.046656\pm0.004351$\\
$\Delta_{\rm loss}^{H}$ & +0.046105 & +0.044806 & +0.041226 & $+0.044046\pm0.002527$\\
$\Delta_{\rm int}$ & +0.001726 & +0.000890 & -0.010446 & $-0.002610\pm0.006799$\\
\bottomrule
\end{tabular}
\end{table}

\section{Full Factorial Recall@$K$}

The following tables are obtained only by applying the registered aggregation to the archived official-test Q/G JSON cells: three Q/G values are averaged within training seed, then mean $\pm$ sample SD is taken across the three training seeds. No model is retrained and no value is inferred from mAP summaries.

\begin{table}[h]
\centering
\caption{CUB factorial Recall@$K$ (query-micro), mean $\pm$ sample SD over training seeds.}
\small
\begin{tabular}{llrrrr}
\toprule
Arm & Level & R@1 & R@5 & R@10 & R@20\\
\midrule
$E_R$ & Class & $0.7718\pm0.0025$ & $0.9355\pm0.0057$ & $0.9656\pm0.0049$ & $0.9799\pm0.0031$\\
$E_R$ & Genus & $0.2417\pm0.0066$ & $0.6462\pm0.0104$ & $0.8134\pm0.0072$ & $0.9417\pm0.0066$\\
$E_R$ & Family & $0.0829\pm0.0012$ & $0.2292\pm0.0065$ & $0.3329\pm0.0051$ & $0.5128\pm0.0090$\\
$H_R$ & Class & $0.7895\pm0.0027$ & $0.9387\pm0.0019$ & $0.9650\pm0.0033$ & $0.9793\pm0.0014$\\
$H_R$ & Genus & $0.2328\pm0.0011$ & $0.6476\pm0.0141$ & $0.8170\pm0.0176$ & $0.9561\pm0.0014$\\
$H_R$ & Family & $0.0698\pm0.0046$ & $0.1971\pm0.0103$ & $0.2901\pm0.0025$ & $0.4559\pm0.0087$\\
$E_S$ & Class & $0.6595\pm0.0063$ & $0.9102\pm0.0027$ & $0.9561\pm0.0012$ & $0.9771\pm0.0006$\\
$E_S$ & Genus & $0.3820\pm0.0095$ & $0.8134\pm0.0043$ & $0.9096\pm0.0038$ & $0.9706\pm0.0022$\\
$E_S$ & Family & $0.1219\pm0.0046$ & $0.3029\pm0.0058$ & $0.4253\pm0.0056$ & $0.6111\pm0.0042$\\
$H_S$ & Class & $0.6006\pm0.0044$ & $0.8803\pm0.0015$ & $0.9439\pm0.0008$ & $0.9737\pm0.0019$\\
$H_S$ & Genus & $0.4052\pm0.0176$ & $0.8370\pm0.0093$ & $0.9366\pm0.0018$ & $0.9734\pm0.0020$\\
$H_S$ & Family & $0.1748\pm0.0136$ & $0.4055\pm0.0039$ & $0.5287\pm0.0080$ & $0.6760\pm0.0076$\\
\bottomrule
\end{tabular}
\end{table}

\begin{table}[h]
\centering
\caption{NABirds Parent-disjoint factorial Recall@$K$ (group-macro), mean $\pm$ sample SD over training seeds.}
\small
\begin{tabular}{llrrrr}
\toprule
Arm & Level & R@1 & R@5 & R@10 & R@20\\
\midrule
$E_R$ & Class & $0.8510\pm0.0070$ & $0.9545\pm0.0059$ & $0.9738\pm0.0050$ & $0.9867\pm0.0035$\\
$E_R$ & Parent & $0.0652\pm0.0098$ & $0.2079\pm0.0083$ & $0.3360\pm0.0126$ & $0.5090\pm0.0188$\\
$E_R$ & Supergroup & $0.0848\pm0.0086$ & $0.2561\pm0.0052$ & $0.3602\pm0.0151$ & $0.5410\pm0.0199$\\
$H_R$ & Class & $0.8507\pm0.0027$ & $0.9537\pm0.0053$ & $0.9762\pm0.0012$ & $0.9874\pm0.0017$\\
$H_R$ & Parent & $0.0615\pm0.0029$ & $0.2127\pm0.0016$ & $0.3348\pm0.0112$ & $0.5119\pm0.0081$\\
$H_R$ & Supergroup & $0.0875\pm0.0087$ & $0.2528\pm0.0240$ & $0.3662\pm0.0199$ & $0.5339\pm0.0158$\\
$E_S$ & Class & $0.6444\pm0.0064$ & $0.8587\pm0.0033$ & $0.9206\pm0.0022$ & $0.9553\pm0.0004$\\
$E_S$ & Parent & $0.1478\pm0.0021$ & $0.4338\pm0.0066$ & $0.6090\pm0.0095$ & $0.7965\pm0.0171$\\
$E_S$ & Supergroup & $0.2324\pm0.0017$ & $0.5603\pm0.0125$ & $0.6959\pm0.0120$ & $0.8148\pm0.0101$\\
$H_S$ & Class & $0.6064\pm0.0073$ & $0.8397\pm0.0155$ & $0.9096\pm0.0087$ & $0.9474\pm0.0004$\\
$H_S$ & Parent & $0.1647\pm0.0096$ & $0.4609\pm0.0160$ & $0.6287\pm0.0110$ & $0.7978\pm0.0182$\\
$H_S$ & Supergroup & $0.2643\pm0.0038$ & $0.5791\pm0.0046$ & $0.7169\pm0.0224$ & $0.8210\pm0.0136$\\
\bottomrule
\end{tabular}
\end{table}

The accepted factorial audits contain complete Recall@$K$ cells for the four factorial arms. The currently available accepted archive does not contain complete R@5/R@10/R@20 raw cells for every historical contextual baseline, so we do not synthesize those values; the main paper's Supplement promise is correspondingly limited to \emph{factorial} Recall@$K$.

\section{Semantic-Alignment Control}

This experiment is separate from the fixed-100-epoch factorial. All arms use the same frozen DINOv2 features and 768--256--32 Euclidean projector, batch size 128, learning rate $10^{-4}$, weight decay $10^{-4}$, dropout 0.1, and temperature $\tau=0.30$. The supervision weights are Leaf-only $(0.25,0,0)$, Leaf+Parent $(0.25,0.50,0)$, and Full/Shuffle $(0.25,0.50,1.00)$. Each arm selects its duration on development data without official-test access; the frozen epochs are 5, 47, 119, and 2 respectively.

\begin{table}[h]
\centering
\caption{NABirds Parent-disjoint semantic-alignment control. Values are mean $\pm$ sample SD over training seeds after Q/G averaging within seed.}
\small
\begin{tabular}{lrrrrr}
\toprule
Arm & Class R@1 & Class mAP & Parent mAP & Supergroup mAP & Mean hierarchy\\
\midrule
Leaf-only & $.8447\pm.0013$ & $.6717\pm.0067$ & $.1901\pm.0013$ & $.4231\pm.0025$ & $.3066\pm.0019$\\
Leaf+Parent & $.8096\pm.0018$ & $.6178\pm.0072$ & $.2441\pm.0072$ & $.4407\pm.0059$ & $.3424\pm.0015$\\
Full taxonomy & $.7478\pm.0078$ & $.5459\pm.0030$ & $.2408\pm.0056$ & $.5391\pm.0024$ & $.3899\pm.0031$\\
Shuffled taxonomy & $.8126\pm.0043$ & $.5874\pm.0034$ & $.1559\pm.0105$ & $.2881\pm.0113$ & $.2220\pm.0107$\\
\bottomrule
\end{tabular}
\end{table}

The structure-preserving shuffle uses fixed seed 20260818, leaves leaf labels unchanged, preserves the multiset of classes per Parent and Parents per Supergroup, has no class-to-Parent fixed points, and necessarily retains 15 Parent-to-Supergroup fixed points under the group-size constraints. Full taxonomy minus shuffled taxonomy is $+0.0849$ Parent mAP, $+0.2510$ Supergroup mAP, and $+0.1679$ mean hierarchy mAP; all three paired training-seed directions are positive for these three contrasts. Leaf+Parent minus Leaf-only mean hierarchy is $+0.0358$ with all three seed directions positive; Full minus Leaf+Parent mean hierarchy is $+0.0475$, again positive for all three seeds, while Parent mAP changes by about $-0.0034$ and improves in only one of three seeds.

\paragraph{Archive limitation.}
The accepted manuscript record contains the frozen means, sample SDs, epochs, shuffle construction, and paired-direction counts above. The exact per-seed semantic-ablation rows are not present in the currently available accepted audit ZIPs, so they are intentionally not reconstructed from means and standard deviations.

\section{Curvature and Radius Mechanism Boundary}

The mechanism study uses the NABirds Parent-disjoint matched-regression setting and treats the training seed as the statistical unit after averaging three Q/G perturbations. Four new hyperbolic configurations are trained for 84 epochs and compared with the previously accepted matched $c=0.1,r=0.4$ and Euclidean controls.

\begin{table}[h]
\centering
\caption{NABirds Parent-disjoint curvature/radius mechanism control. Mean $\pm$ sample SD over training seeds.}
\small
\begin{tabular}{lrrr}
\toprule
Configuration & Parent mAP & Supergroup mAP & Mean hierarchy\\
\midrule
Near-flat $c=10^{-6},r=.4$ & $.196609\pm.001516$ & $.489188\pm.000151$ & $.342898\pm.000682$\\
Base $c=.1,r=.4$ & $.196308\pm.001256$ & $.488650\pm.000356$ & $.342479\pm.000457$\\
Strong curvature $c=1,r=.4$ & $.195269\pm.000472$ & $.485061\pm.000216$ & $.340165\pm.000173$\\
Small radius $c=.1,r=.2$ & $.197770\pm.001646$ & $.471127\pm.003440$ & $.334449\pm.001068$\\
Large radius $c=.1,r=.8$ & $.199846\pm.004081$ & $.478574\pm.009418$ & $.339210\pm.006642$\\
Matched Euclidean & $.198735\pm.003163$ & $.482696\pm.004045$ & $.340715\pm.003594$\\
\bottomrule
\end{tabular}
\end{table}

At fixed $r=.4$, Base minus Near-flat mean hierarchy is $-0.000419\pm0.000229$ and is negative for all three training seeds. Strong-curvature minus Base is $-0.002314\pm0.000515$, again with a consistent negative direction. Small-radius minus Base is $-0.008031\pm0.000622$ and decreases in all three seeds. Large-radius minus Base averages $-0.003269\pm0.006595$ with only one of three seeds negative, illustrating a non-monotonic level trade-off. Near-flat minus matched Euclidean is $+0.002183\pm0.003793$ and is positive in only two seeds; because that comparison also changes normalization, exponential mapping, and the radius constraint, it is not treated as a pure-curvature intervention.

These results are used only as a \emph{mechanism boundary}: they do not support attributing the small matched geometry differences to stronger negative curvature. Exact per-seed mechanism rows are not present in the currently available accepted audit archive and are therefore not reconstructed.

\section{Harmonized CUB Group-Macro Reporting Sensitivity}

After the CUB factorial was frozen, the 12 already frozen projected-test arrays were re-scored with group-macro aggregation using the NABirds aggregation family. No training, hyperparameter selection, or checkpoint selection was performed.

\begin{table}[h]
\centering
\caption{Post-hoc CUB group-macro mean-hierarchy effects.}
\small
\begin{tabular}{lrrrr}
\toprule
Effect & 42 & 2024 & 3407 & Mean $\pm$ SD\\
\midrule
$\Delta_{\rm geom}^{R}$ & +0.008632 & +0.008104 & +0.009976 & $+0.008904\pm0.000965$\\
$\Delta_{\rm geom}^{S}$ & +0.004214 & +0.004807 & +0.004921 & $+0.004647\pm0.000380$\\
$\Delta_{\rm loss}^{E}$ & +0.061003 & +0.062655 & +0.062948 & $+0.062202\pm0.001048$\\
$\Delta_{\rm loss}^{H}$ & +0.056585 & +0.059358 & +0.057892 & $+0.057945\pm0.001387$\\
$\Delta_{\rm int}$ & -0.004418 & -0.003296 & -0.005055 & $-0.004256\pm0.000890$\\
\bottomrule
\end{tabular}
\end{table}

The reporting audit verified 12/12 frozen projected arrays and 36/36 Q/G cells. Recomputed query-micro metrics reconcile to the accepted primary results with maximum absolute mAP delta 0.00000679, below the frozen $10^{-5}$ tolerance. This sensitivity preserves the qualitative scale separation between loss and geometry but is explicitly post-hoc and does not replace CUB's registered query-micro endpoint.

\section{Fairness, Leakage, and Reproducibility Audit}

\begin{table}[h]
\centering
\caption{Final factorial fairness and leakage checks from the accepted pre-test/final audits.}
\small
\begin{tabular}{lcc}
\toprule
Check & CUB & NABirds Parent-disjoint\\
\midrule
Training samples & 5,994 & 19,021\\
Trainable parameters/cell & 205,600 & 205,600\\
Batch size & 128 & 128\\
Epochs/cell & 100 & 100\\
Updates/epoch & 46 & 148\\
Training seeds & 42/2024/3407 & 42/2024/3407\\
Q/G seeds & 42/2024/3407 & 42/2024/3407\\
Final training runs & 12/12 verified & 12/12 verified\\
Official Q/G cells & 36/36 verified & 36/36 verified\\
Same trainable init within seed & pass & pass\\
Same batch order within seed & pass & pass\\
Same parameter-update budget & pass & pass\\
Checkpoint reload max abs. diff. & 0 & 0\\
Temperature-tune checkpoints reused & no & no\\
All final arms retrained from scratch & yes & yes\\
Configuration selection used test & no & no\\
Training finished before test metrics & yes & yes\\
Official test read after Final lock & yes & yes\\
\bottomrule
\end{tabular}
\end{table}

For CUB, the hyperbolic ball radius is 0.7071 and the maximum audited final hyperbolic embedding norm across accepted runs is 0.3622; for NABirds the corresponding values are 3.1623 and 0.3979. Both therefore satisfy the audited ball constraint with substantial margin.

The official-test split hashes, model-state hashes, and lock hashes are retained in the accepted audit archives rather than repeated in the paper text. The final audits verify that official-test metrics were accessed only after the corresponding Final lock and after all 12 training runs were complete.

\section{Archived-Evidence Limits}

For transparency, the following reproducibility details are not fully recoverable from the currently available accepted artifact set:
\begin{itemize}
\item the exact raw per-seed rows of the semantic-alignment ablation;
\item the exact raw per-seed rows of the curvature/radius mechanism study;
\item the realized CUB taxonomy mapping CSV, manual-alias JSON, and external eBird/Clements v2025 source CSV;
\item the exact final Q/G NPZ index arrays, although the project split utilities and accepted seeds/sizes/SHA256 identifiers are now included;
\item the Geoopt package version, complete environment freeze, and library-internal \texttt{projx} boundary epsilon.
\end{itemize}
None of these gaps is filled by back-calculation, version guessing, or regeneration presented as original data. The main factorial, temperature-selection, Q/G Recall@$K$, seed-level effects, and fairness/test-seal claims are supported directly by the accepted JSON/lock audits.